\documentclass[11pt]{article}

\usepackage[preprint]{acl}

\usepackage{times}
\usepackage{latexsym}

\usepackage[T1]{fontenc}

\usepackage[utf8]{inputenc}

\usepackage{microtype}

\usepackage{inconsolata}

\usepackage{graphicx}
\usepackage{enumitem}
\usepackage{amsmath}

\title{Intra-Rubric Reward Divergence: Mitigating Rubric Hacking through Latent Dimension Recovery and Balancing}

\author{
  \textbf{Jianqiang Xia\textsuperscript{1,2,3,\textasteriskcentered}},
  \textbf{Han Xu\textsuperscript{2,\textasteriskcentered \textdagger}},
  \textbf{Dianxi Shi\textsuperscript{1,3,\textdaggerdbl}}
\\
  \textsuperscript{1}Shanghai Jiao Tong University,
  \textsuperscript{2}Baidu, Inc.,
  \textsuperscript{3}Intelligent Game and Decision Lab,
}

\begin{document}
\maketitle
\begingroup
\renewcommand\thefootnote{}
\footnote{\textsuperscript{\textasteriskcentered}Equal contribution. \textsuperscript{\textdagger}Project leader. \textsuperscript{\textdaggerdbl}Corresponding author. Work done at Baidu, Inc.}
\addtocounter{footnote}{-1}
\endgroup
\begin{abstract}
Rubric hacking limits both performance and training robustness in rubric-based reinforcement learning. By decomposing rubrics into sub-dimensions, we reveal the Intra-Rubric Reward Divergence: as the total rubric score increases, certain sub-dimension scores systematically decrease, suggesting the model sacrifices specific quality aspects to optimize the aggregated objective. From the sub-dimension perspective, we observe two phenomena: (1) dimension imbalance, causing the policy to favor heavily covered dimensions; (2) dimension absence, leading to a systematic gap between rewards and true quality. To mitigate these issues, we propose Fair-Race Dimension Aggregation (FRDA), which has generations compete within each dimension separately to mitigate rubric imbalance, and Residual Semantic Dimension Completion (RSDC), which mines evaluation dimensions implicitly encoded in a holistic preference signal to complete rubric coverage. Experiments in chat, science, and medical domains indicate that FRDA mitigates rubric hacking with more stable improvements, while RSDC outperforms both single-reward training and naive multi-reward aggregation, yielding generally higher-quality responses.
\end{abstract}

\section{Introduction}
\label{sec:introduction}

\begin{figure}[t]
  \centering
  \includegraphics[width=\columnwidth]{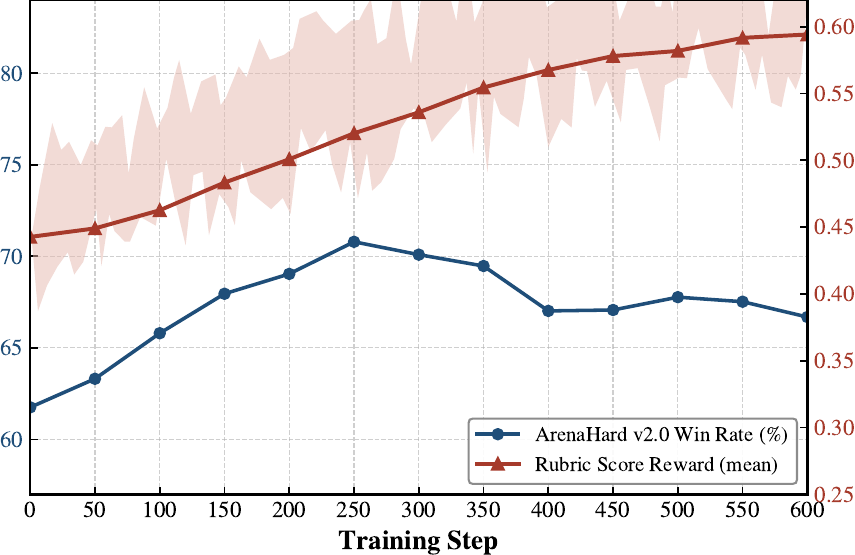}
  \caption{\textbf{Rubric Hacking.} Taking the chat rubric training task as an example: as the overall rubric score keeps rising, the ArenaHard v2.0 win rate declines.}
  \label{fig:figure3}
\end{figure}

\textbf{Rubric Dimension Decomposition.} Rubric-based reinforcement learning~\citep{RaR,RubricAnchors,Checklist,POW3R} extends the verifiable reward reinforcement learning (RLVR) paradigm~\citep{DeepseekR1,DAPO,Reinforce++,GRPO_accurary,GRPO_format,GRPO_length,GDPO} to open-ended tasks with fine-grained, interpretable signals, but rubric hacking remains a key bottleneck limiting its performance and stability. As shown in Figure~\ref{fig:figure3}, rubric hacking~\citep{RubricHacking,CHERRL} typically manifests as a continuous increase in overall rubric reward during training, while generation quality declines after an initial improvement. Existing work~\citep{RubricDropout} mitigates this by reducing overfitting to the judge model's biased criteria, without analyzing how the internal composition of the overall rubric reward changes. To investigate this, we track how reward shifts within the rubric during training. Since prompts vary in criteria count and criteria are overly specific and semantically fragmented, criterion-level analysis is difficult. We therefore decompose rubrics into semantic dimensions per task domain, each abstract, relatively independent, and covering a major aspect of generation quality.

\begin{figure*}[t]
  \centering
  \includegraphics[width=\textwidth]{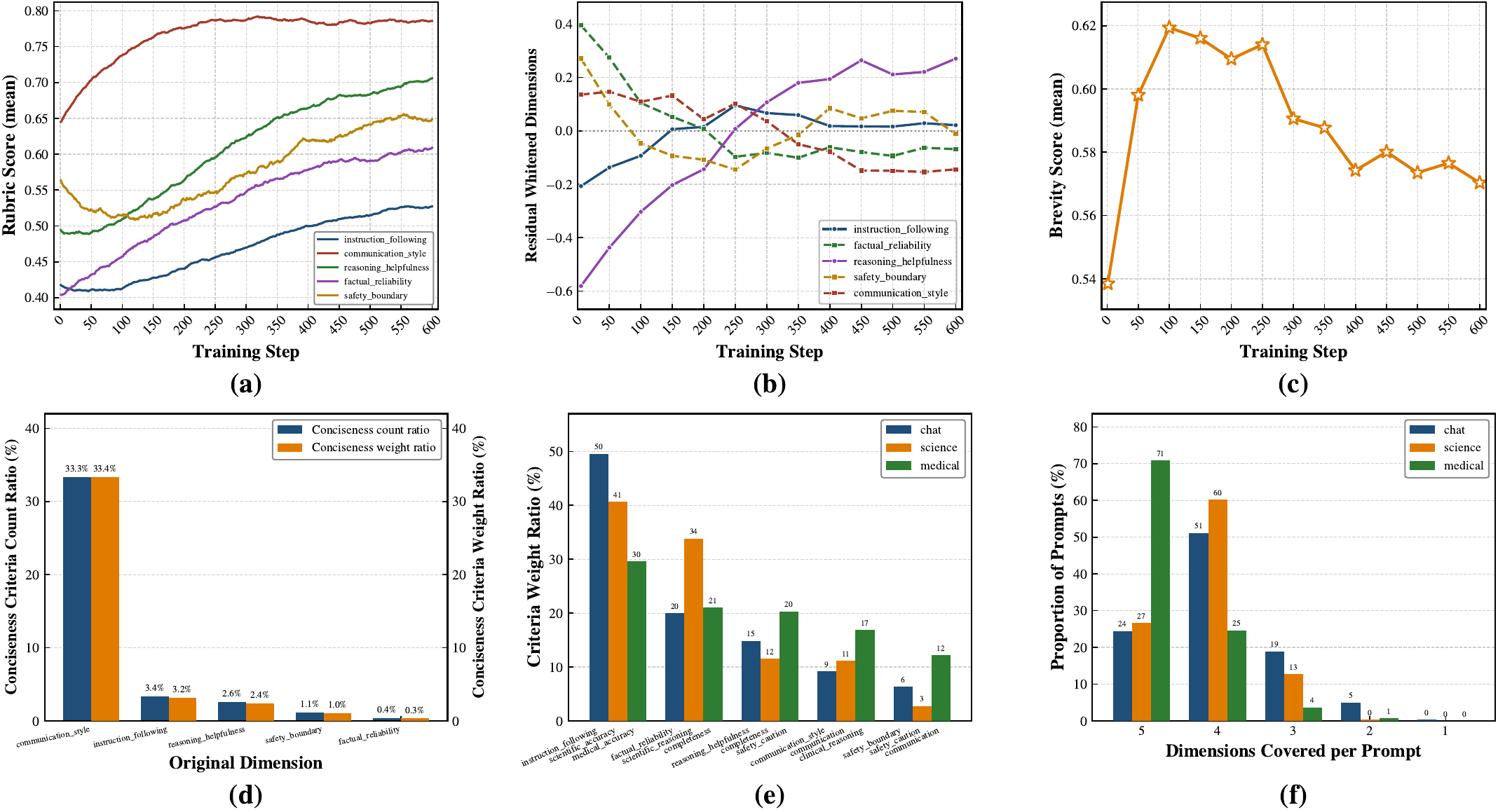}
  \caption{(a) Rubric reward of each semantic dimension, with communication style stagnating. (b) The Cholesky-whitened communication style dimension shows a clear downward trend. (c) Rubric reward of explicitly extracted conciseness criteria declines steadily. (d) Count and weight ratios of conciseness criteria across the five original dimensions. (e) Per-dimension criteria weight ratios across the chat, science, and medical domains, where the medical domain is the most balanced. (f) Distribution of semantic dimensions covered per prompt.}
  \label{fig:figure12}
\end{figure*}

\textbf{Intra-Rubric Reward Divergence.} Taking the chat task as an example, we divide rubric criteria into five explicit semantic dimensions: instruction following, factual reliability, reasoning helpfulness, communication style, and safety boundary. As shown in Figure~\ref{fig:figure12}(a), the overall rubric reward keeps rising while the communication style dimension stagnates. To rule out semantic overlap inevitably introduced by manual partitioning, we apply Cholesky residual whitening~\citep{Cholesky} to decorrelate the explicit semantic dimensions, yielding whitened latent dimensions that each retain only the reward information unique to that dimension. As shown in Figure~\ref{fig:figure12}(b), the latent dimensions of factual reliability and communication style exhibit a clear downward trend. Guided by the communication style latent dimension, we trace back training samples and find that its decline corresponds to the model ignoring finer-grained requirements such as readability and conciseness, instead inflating rubric hit rates by enumerating irrelevant or redundant content---a behavior that runs counter to true generation quality. We further explicitly extract conciseness-related criteria: as shown in Figure~\ref{fig:figure12}(c), their scores also decline steadily, and as shown in Figure~\ref{fig:figure12}(d), they account for 33.37\% of the communication style dimension's weight, closely matching the latent dimension's trend. This cross-validation shows that the pattern is not random fluctuation; rather, as the overall rubric reward rises, the model systematically sacrifices certain sub-dimensions, a phenomenon we term \emph{Intra-Rubric Reward Divergence}.

\textbf{Rubric Dimension Imbalance.} Motivated by this finding, we analyze the distribution of rubric criteria across semantic dimensions in three training domains and observe a clear imbalance, as shown in Figure~\ref{fig:figure12}(e). In the chat task, in both the original and whitened semantic spaces, upward-trending dimensions carry much larger weights than stagnant or declining ones, indicating that densely covered dimensions dominate reward aggregation and cause the model to sacrifice sparsely covered ones. To mitigate this, inspired by the multi-reward RL algorithm GDPO~\citep{GDPO}, we propose Fair-Race Dimension Aggregation (FRDA): first compute rubric sub-dimension group advantages, then aggregate them with equal weights, letting generations compete separately within each dimension. This forces the model to perform well on more dimensions to achieve high reward, weakening the dominance of dense dimensions and reducing sacrifice of sparse dimensions.

\textbf{Rubric Dimension Absence.} As shown in Figure~\ref{fig:figure12}(f), a single prompt covers only 2--5 semantic dimensions, since annotators cannot cover all evaluation criteria---a fundamental difference from RLVR. A holistic preference signal implicitly encodes a relatively complete human evaluation system. To this end, we propose Residual Semantic Dimension Completion (RSDC): leveraging the decorrelation property of Cholesky whitening residuals, it mines from the holistic preference signal semantic dimensions not yet annotated in the rubric, filling the potential coverage gap in rubric criteria. 


Our main contributions are:

\begin{itemize}[leftmargin=*, itemsep=0.5\baselineskip, topsep=0.3\baselineskip, parsep=0pt]
\item We identify \textbf{intra-rubric reward divergence}, a previously underexplored form of rubric hacking in which the aggregate rubric reward increases while specific semantic dimensions systematically decline.
\item We propose \textbf{Fair-Race Dimension Aggregation (FRDA)}, which has generations compete within each dimension separately to mitigate rubric imbalance.
\item We propose \textbf{Residual Semantic Dimension Completion (RSDC)}, the first to implicitly mine complementary quality dimensions from global preferences without explicitly adding rubric criteria.
\item Extensive experiments on chat, science, and medical tasks show that FRDA effectively mitigates rubric hacking, and RSDC significantly outperforms both single-reward training and naive multi-reward aggregation, offering a new perspective on jointly leveraging interpretable rubric rewards and holistic preference rewards.
\end{itemize}

\section{Related Work}
\label{sec:related_work}

\noindent\textbf{Rubric Reinforcement Learning.} RaR~\citep{RaR} and Rubric Anchors~\citep{RubricAnchors} established rubric scores as fine-grained reward signals for reinforcement learning on open-ended tasks, while Checklist Feedback~\citep{Checklist} uses instruction-specific checklists to provide training rewards. OpenRubric~\citep{OpenRubric} and RubricHub scale rubric construction through LLM-based generation. These works directly use the overall rubric score as the training reward. POW3R~\citep{POW3R} reweights criteria according to the variance of judgments within a sampled group, concentrating optimization on criteria with greater discriminative power.

\noindent\textbf{Rubric Hacking.} Mahmoud et al.~\citep{RubricHacking} document rubric reward hacking as divergence between the training verifier and a stronger judge, attributing it to verifier failure and rubric limitations; CHERRL~\citep{CHERRL} reproduces this by injecting judge biases and detecting their exploitation. Both are diagnostic, revealing rubric hacking but offering no solutions. The closest work, Rubric Dropout~\citep{RubricDropout}, inspired by neural dropout, randomly drops rubric parts so the model cannot rely on any particular criterion to game the rubric. However, existing methods ignore internal dynamics within the rising overall rubric reward. Revealing Intra-Rubric Reward Divergence via rubric dimension-wise decomposition is our core insight; we thus propose Fair-Race Dimension Aggregation (FRDA) to mitigate rubric hacking, unlike existing methods. This offers a new perspective; our method is orthogonal to Dropout, and their combination is future work.


\section{Method}
\label{sec:method}

In this section, we first introduce the rubric reward and the semantic-dimension decomposition (Section~\ref{sec:method_prelim}). We then present Fair-Race Dimension Aggregation (FRDA) (Section~\ref{sec:method_balanced}) and Residual Semantic Dimension Completion (RSDC) (Section~\ref{sec:method_latent}).

\begin{figure*}[t]
  \centering
  \includegraphics[width=\textwidth]{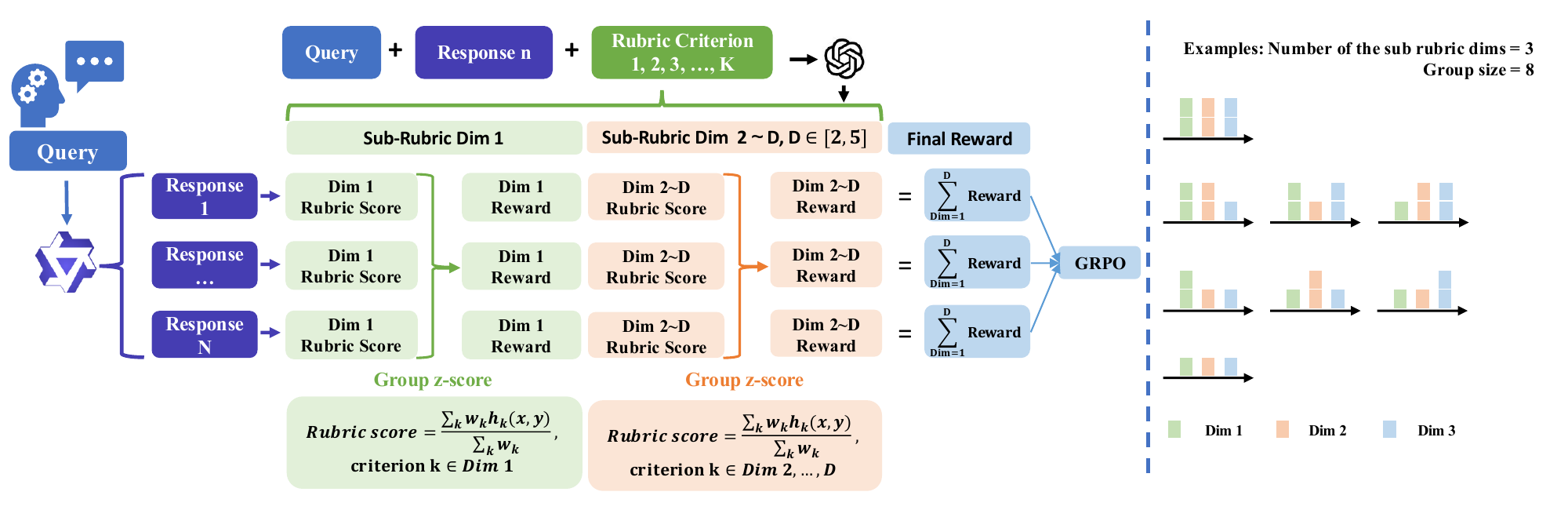}
  \caption{Overview of Fair-Race Dimension Aggregation (FRDA). Each covered dimension is first z-scored within the sampled group, then aggregated with equal weights, letting generations compete fairly within each dimension regardless of its criterion count or weight.} 
  \label{fig:method1}
\end{figure*}

\subsection{Preliminaries}
\label{sec:method_prelim}

\noindent\textbf{Rubric Reward.} Let $\pi_\theta$ be the policy and $x$ a prompt from the training set. Each prompt is paired with a rubric $\mathcal{C}(x)=\{(c_k, w_k)\}_{k=1}^{K_x}$, where $c_k$ is a natural-language criterion and $w_k>0$ is its importance weight. Given a response $y$, an LLM judge $J$ decides whether each criterion is satisfied, yielding a binary verdict $h_k(x,y)=J(x,y,c_k)\in\{0,1\}$; in practice, all criteria of a prompt are judged in a single call. The overall rubric reward~\citep{RaR,rubrichub} is the weight-normalized hit rate:
\begin{equation}
  \label{eq:rubric_reward}
  r(x,y) = \frac{\sum_{k=1}^{K_x} w_k\, h_k(x,y)}{\sum_{k=1}^{K_x} w_k} \in [0,1].
\end{equation}

\noindent\textbf{Semantic Dimension Decomposition.} Rubric criteria are numerous, overly specific, and semantically fragmented, which hinders criterion-level analysis. We therefore group them into $D$ semantic dimensions per domain under three principles: (i) \emph{abstraction}, each dimension summarizes a family of related criteria; (ii) \emph{relative independence}, each dimension captures a relatively independent aspect of generation quality; and (iii) \emph{completeness}, the dimensions jointly cover the rubric so that every criterion belongs to exactly one dimension. We use GPT-5.4~\citep{gpt5.4} to classify each criterion, yielding a mapping $\phi$ with $\phi(c_k)\in\{1,\dots,D\}$. Taking chat as an example, the $D=5$ dimensions are instruction following, factual reliability, reasoning helpfulness, communication style, and safety boundary. Hereafter, we refer to dimensions by index, i.e., dimension $1,\dots,D$.

\subsection{Fair-Race Dimension Aggregation (FRDA)}
\label{sec:method_balanced}

\noindent\textbf{Imbalance in Rubric Aggregation.} Let $\mathcal{K}_d(x)=\{k : \phi(c_k)=d\}$ be the indices of criteria assigned to dimension $d$, $\mathcal{S}(x)=\{d : \mathcal{K}_d(x)\neq\emptyset\}$ the dimensions covered by $x$, and $W_d(x)=\sum_{k\in\mathcal{K}_d(x)} w_k$. Reusing the judge verdicts, the dimension-level reward is
\begin{equation}
  \label{eq:dim_reward}
  r_d(x,y) = \frac{\sum_{k\in\mathcal{K}_d(x)} w_k\, h_k(x,y)}{W_d(x)}, \quad d\in\mathcal{S}(x),
\end{equation}
and Eq.~\eqref{eq:rubric_reward} is exactly their weighted mean:
\begin{equation}
  \label{eq:rubric_decomp}
  r(x,y) = \sum_{d\in\mathcal{S}(x)} \frac{W_d(x)}{\sum_{d'\in\mathcal{S}(x)} W_{d'}(x)}\, r_d(x,y).
\end{equation}
Standard Group Relative Policy Optimization (GRPO) normalizes $r(x,y_i)$ within a group of $G$ responses $\{y_i\}_{i=1}^{G}$ sampled from $\pi_{\theta_\text{old}}$ to obtain the advantage. By Eq.~\eqref{eq:rubric_decomp}, dimensions with dense, heavily weighted criteria dominate this signal, so the policy can raise $r$ while sacrificing sparsely covered dimensions. Inspired by GDPO, we instead normalize each dimension within the group before aggregation.

\noindent\textbf{Dimension-Wise Normalization and Aggregation.} With $r_{i,d}=r_d(x,y_i)$, each covered dimension is normalized within the group:
\begin{equation}
  \label{eq:dim_adv}
  \hat{A}_{i,d} = \frac{r_{i,d} - \mu_d}{\sigma_d + \epsilon},
\end{equation}
where $\mu_d$ and $\sigma_d$ are the mean and standard deviation of $\{r_{j,d}\}_{j=1}^{G}$ and $\epsilon$ is a small constant. The normalized advantages are then averaged with equal weights:
\begin{equation}
  \label{eq:balanced_agg}
  R_i = \frac{1}{|\mathcal{S}(x)|} \sum_{d\in\mathcal{S}(x)} \hat{A}_{i,d}.
\end{equation}
Unlike GDPO, whose reward set is shared by all prompts, the covered dimensions vary across prompts (Figure~\ref{fig:figure12}(f)), so we average only over $\mathcal{S}(x)$. Finally, $R_i$ is normalized again within the group to keep the advantage scale consistent across prompts:
\begin{equation}
  \label{eq:balanced_final}
  A_i = \frac{R_i - \operatorname{mean}\big(\{R_j\}_{j=1}^{G}\big)}{\operatorname{std}\big(\{R_j\}_{j=1}^{G}\big) + \epsilon},
\end{equation}
which replaces the standard GRPO advantage in the clipped surrogate objective.

\noindent\textbf{Why It Balances.} After Eq.~\eqref{eq:dim_adv}, every covered dimension has zero mean and unit variance within the group regardless of its criterion count or weights, so it contributes equally to Eq.~\eqref{eq:balanced_agg} rather than in proportion to $W_d(x)$ as in Eq.~\eqref{eq:rubric_decomp}. A response thus needs to outperform its peers on more covered dimensions to obtain a high advantage.

\subsection{Residual Semantic Dimension Completion (RSDC)}
\label{sec:method_latent}

\begin{figure}[t]
  \centering
  \includegraphics[width=\columnwidth]{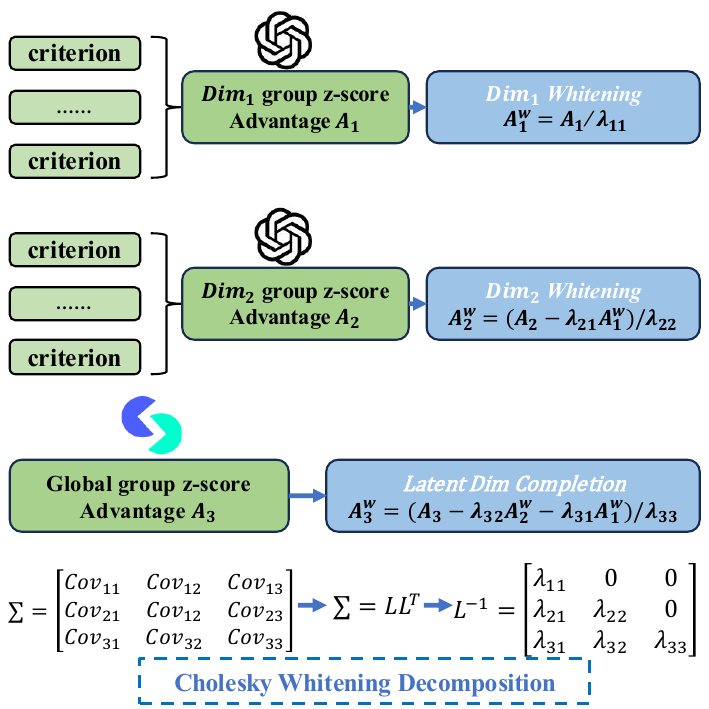}
  \caption{Illustration of Residual Semantic Dimension Completion (RSDC).} 
  \label{fig:method2}
\end{figure}

Balanced aggregation cannot recover dimensions that a rubric misses (Figure~\ref{fig:figure12}(f)). A global preference reward model implicitly encodes a more complete evaluation system, but its signal may largely overlap with the rubric dimensions, and adding it directly would reward covered aspects twice. We use Cholesky whitening to extract only its complementary component.

\noindent\textbf{Cholesky Whitening as Residual Whitening.} For a zero-mean vector $\mathbf{v}=(v_1,\dots,v_n)^\top$ with positive definite covariance $\Sigma=LL^\top$, where $L$ is lower-triangular, $\mathbf{u}=L^{-1}\mathbf{v}$ has identity covariance. Since $L^{-1}$ is also lower-triangular, each $u_p$ is the standardized residual of linearly regressing $v_p$ on its predecessors $\mathbf{v}_{<p}$:
\begin{equation}
  \label{eq:residual_whiten}
  u_p = \frac{v_p - \boldsymbol{\beta}_p^\top \mathbf{v}_{<p}}{\tau_p},
\end{equation}
where $\boldsymbol{\beta}_p=\Sigma_{<p,<p}^{-1}\Sigma_{<p,p}$ and $\tau_p^2=\Sigma_{pp}-\Sigma_{p,<p}\boldsymbol{\beta}_p$. Unlike principal component analysis (PCA) or zero-phase component analysis (ZCA) whitening, which mix all coordinates, Cholesky whitening keeps in each later coordinate only the information that earlier ones cannot explain. This property suits our setting well. Placing the global preference last strips its linear overlap with the rubric dimensions, so its whitened coordinate forms a latent dimension complementary to the rubric, while the whitened rubric dimensions are unaffected by the preference reward.

\noindent\textbf{Latent Dimension Completion.} Let $r_\text{pref}(x,y)$ be the score of a Bradley--Terry reward model (Skywork-Reward-V2-Llama-3.1-8B in our experiments).
We treat it as dimension $D{+}1$, covered by every prompt, set $\tilde{\mathcal{S}}(x)=\mathcal{S}(x)\cup\{D{+}1\}$ and $r_{i,D{+}1}=r_\text{pref}(x,y_i)$, and group-normalize it by Eq.~\eqref{eq:dim_adv} like the rubric dimensions. For response $y_i$, let $\hat{\mathbf{a}}_i=(\hat{A}_{i,d})_{d\in\tilde{\mathcal{S}}(x)}$ with the covered rubric dimensions first and dimension $D{+}1$ last, and let $\Sigma_{\tilde{\mathcal{S}}(x)}$ be the corresponding principal submatrix of the covariance $\Sigma$ of all $D{+}1$ dimension advantages. Whitening is performed in the covered subspace:
\begin{equation}
  \label{eq:subspace_whiten}
  \Sigma_{\tilde{\mathcal{S}}(x)} = L_x L_x^\top, \qquad \tilde{\mathbf{a}}_i = L_x^{-1}\hat{\mathbf{a}}_i,
\end{equation}
where $\tilde{\mathbf{a}}_i=(\tilde{A}_{i,d})_{d\in\tilde{\mathcal{S}}(x)}$. By Eq.~\eqref{eq:residual_whiten}, the latent dimension $\tilde{A}_{i,D{+}1}$ is the standardized residual of $\hat{A}_{i,D{+}1}$ after removing the part linearly explained by the covered rubric dimensions. Components related to uncovered dimensions are retained, which fills the coverage gap of each prompt. The whitened advantages are aggregated with equal weights, replacing Eq.~\eqref{eq:balanced_agg},
\begin{equation}
  \label{eq:latent_agg}
  R_i = \frac{1}{|\tilde{\mathcal{S}}(x)|} \sum_{d\in\tilde{\mathcal{S}}(x)} \tilde{A}_{i,d},
\end{equation}
and converted to the final advantage by Eq.~\eqref{eq:balanced_final}.



\begin{figure*}[t]
  \centering
  \includegraphics[width=\textwidth]{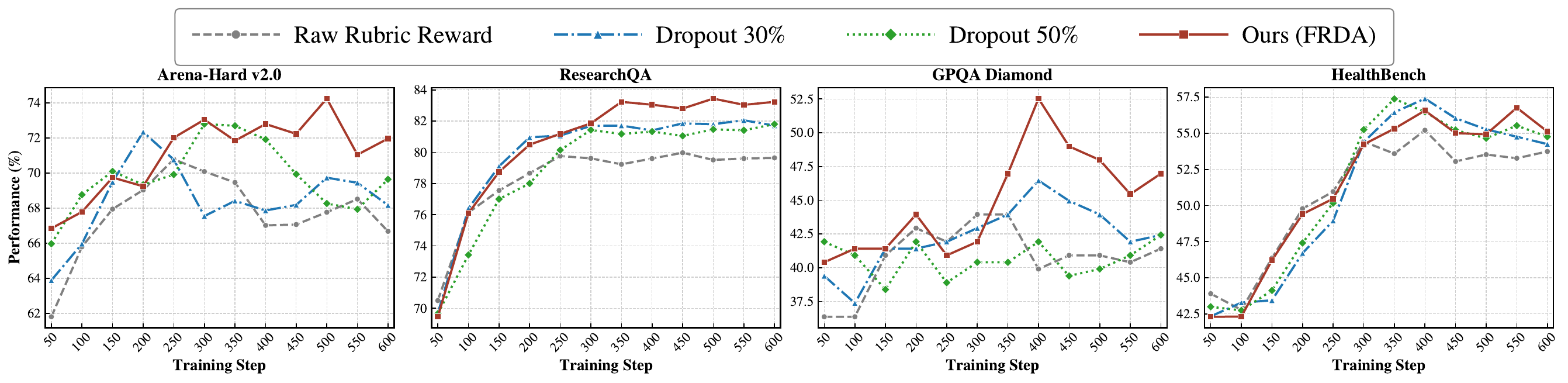}
  \caption{
    Qwen3-4B performance across training steps 50--600 on four
    benchmarks. FRDA outperforms raw rubric reward across domains,
    with larger gains over Dropout on Chat and Science.
  }
  \label{fig:figure31}
\end{figure*}

\begin{table*}[t]
  \centering
  \resizebox{\textwidth}{!}{%
  \setlength{\tabcolsep}{4pt}
  \begin{tabular}{l*{4}{ccc}}
    \hline
    & \multicolumn{3}{c}{\textbf{Arena-Hard v2.0}}
    & \multicolumn{3}{c}{\textbf{ResearchQA}}
    & \multicolumn{3}{c}{\textbf{GPQA Diamond}}
    & \multicolumn{3}{c}{\textbf{HealthBench}} \\
    \textbf{Method}
    & Peak & w-mean & $\Delta$
    & Peak & w-mean & $\Delta$
    & Peak & w-mean & $\Delta$
    & Peak & w-mean & $\Delta$ \\
    \hline
    Qwen3-4B
    & 61.75 & 61.75 & --
    & 60.65 & 60.65 & --
    & 36.36 & 36.36 & --
    & 34.59 & 34.59 & -- \\
    Raw Rubric Reward
    & 70.79 & 67.41 & --
    & 79.97 & 79.56 & --
    & 43.94 & 40.71 & --
    & 55.21 & 53.76 & -- \\
    Dropout 30\%
    & 72.35 & 68.68 & +1.27
    & 82.05 & 81.76 & +2.21
    & 46.46 & 43.94 & +3.23
    & \textbf{57.40} & 55.54 & +1.78 \\
    Dropout 50\%
    & 72.70 & 69.54 & +2.13
    & 81.80 & 81.41 & +1.86
    & 47.47 & 44.95 & +4.24
    & 57.38 & 55.33 & +1.57 \\
    FRDA (Ours)
    & \textbf{74.25} & \textbf{72.46} & \textbf{+5.05}
    & \textbf{83.47} & \textbf{83.16} & \textbf{+3.60}
    & \textbf{55.56} & \textbf{49.70} & \textbf{+8.99}
    & 56.77 & \textbf{55.68} & \textbf{+1.92} \\
    \hline
  \end{tabular}%
  }
  \caption{
    Qwen3-4B results on four benchmarks.
    Peak is the best score across evaluated checkpoints from
    steps 50--600; w-mean averages evaluated checkpoints from
    steps 400--600.
    $\Delta$ is the w-mean gain over raw rubric reward, in points.
    Bold indicates the best result in each column.
  }
  \label{tab:main_results}
\end{table*}

\section{Experiments}
\label{sec:experiment}
We validate the effectiveness of FRDA across the Chat, Science, and Medical domains, and validate RSDC on the Chat domain, where the holistic preference reward signal is readily available.

\subsection{Setup}
\label{sec:experiment_setup}

\noindent\textbf{Models and Training Data.}
We use Qwen3-4B~\citep{qwen3} as the policy model and
GPT-4.1 mini~\citep{gpt41mini_snapshot} as the
rubric judge during training.
We sample 6{,}400 training prompts from each of the Chat,
Science, and Medical subsets of RubricHub~\citep{rubrichub}.
For RSDC on Chat, we additionally use
Skywork-Reward-V2-Llama-3.1-8B~\cite{skyworker}
to provide the holistic preference reward $r_{\mathrm{pref}}$.

\noindent\textbf{Evaluation.}
We evaluate on Arena-Hard v2.0~\cite{areanahard} for Chat,
ResearchQA~\cite{researchqa} and
GPQA Diamond~\cite{gpqa} for Science, and
HealthBench~\cite{healthbench} for Medical.
We check for overlap between RubricHub Science subsets and ResearchQA
and remove overlapping samples before evaluation.
We evaluate on 500 prompts randomly sampled ResearchQA prompts and 1{,}000 uniformly sampled HealthBench prompts.
For each benchmark, we follow its recommended evaluation
settings.

\noindent\textbf{Training and Baselines.}
We train for 600 steps with a learning rate of $10^{-6}$,
a batch size of 32, and 8 rollouts per prompt.
Our main baselines are raw rubric reward, which applies GRPO
to the aggregate rubric score, and Rubric Dropout~\citep{RubricDropout}, which
randomly removes a fraction of rubric criteria when computing
the reward.
We directly adopt the best-practice dropout rates of 30\% and 50\% from the Rubric Dropout~\citep{RubricDropout}.
Further ablations compare different dimension aggregation
and rubric--preference fusion strategies.

\noindent\textbf{Metrics.}
To capture both peak performance and late-training degradation, we report
\emph{Peak}, the best test score over checkpoints from steps 50--600, and
\emph{w-mean}, the mean test score over checkpoints from steps 400--600.
Similar to Rubric Dropout, we use steps 400--600 to probe degradation, as
rubric hacking typically emerges in the last third of training.
$\Delta$ denotes the w-mean gain over the raw rubric reward baseline on
the same benchmark.

\subsection{Effectiveness of Fair-Race Dimension Aggregation}
\label{sec:experiment_balanced}

As shown in Table~\ref{tab:main_results}, on Chat and Science, FRDA improves over raw rubric reward in w-mean by $\Delta=+5.05$ on Arena-Hard v2.0, +3.60 on ResearchQA, and +8.99 on GPQA Diamond, exceeding the better Dropout variant on each by +2.92, +1.40, and +4.75 points. On Medical (HealthBench), FRDA still improves over raw rubric reward ($\Delta=+1.92$) and performs on par with Dropout. As Figure~\ref{fig:figure12}(e) shows, the Chat and Science training sets have a more imbalanced per-dimension criteria weight distribution than Medical. This is consistent with the results above: where training-time dimension weights are more skewed, computing advantages within each dimension and then aggregating them with equal weights can effectively mitigate dimension imbalance, in line with FRDA's larger margin over Dropout's generic overfitting regularization; on the comparatively balanced Medical set, there is less imbalance for FRDA's structural adjustment to act on, in line with its performance being close to Dropout while still improving over raw rubric reward. These results suggest FRDA and Dropout address rubric hacking from two different, orthogonal angles, one structural and one regularization-based: Dropout provides consistently stable optimization, whereas FRDA is more effective when the training set has imbalanced dimension coverage. Combining the two could be worth exploring in future work.

\subsection{FRDA Analysis and Ablation}
\label{sec:experiment_ablation}

\begin{figure}[t]
  \centering
  \includegraphics[width=\columnwidth]{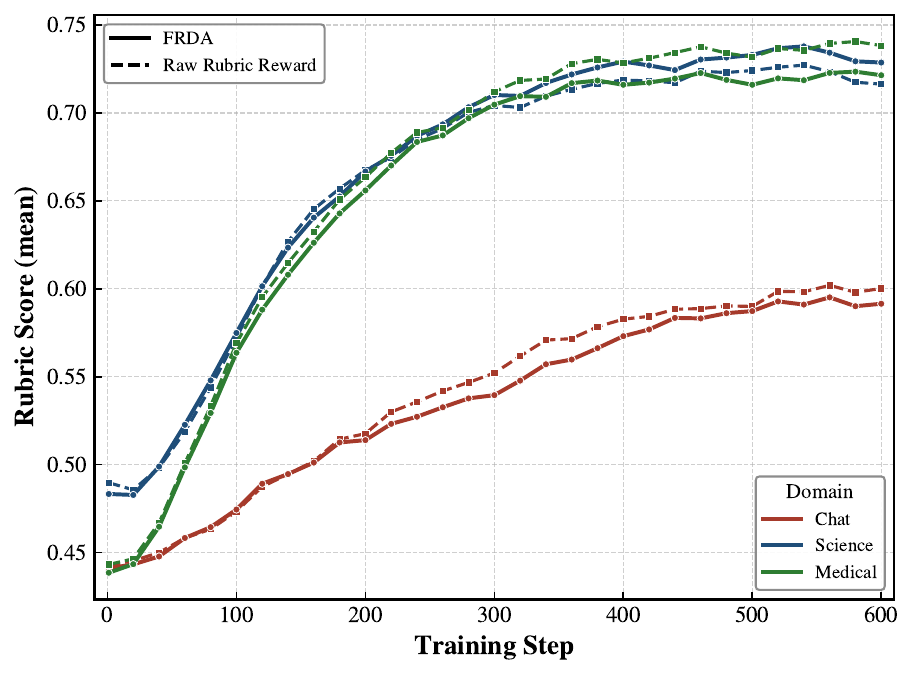}
  \caption{
    Aggregate rubric reward during Qwen3-4B training on Chat,
    Science, and Medical.
    FRDA and raw rubric reward achieve similar aggregate rewards.
  }
  \label{fig:figure32}
\end{figure}

\noindent\textbf{FRDA Mitigates Rubric Hacking Rather Than Merely Slowing Training.} One might attribute FRDA's gains (Figure~\ref{fig:figure31}, Table~\ref{tab:main_results}) to slower training, such that the policy has not yet reached the stage of sacrificing sparsely covered dimensions. Figure~\ref{fig:figure32} rules this out: in all three domains, FRDA's aggregate rubric reward closely tracks that of raw rubric reward at nearly every step, with no delay. Yet FRDA reaches a higher test peak and stays above raw rubric reward over steps 400--600. The gain thus stems not from how fast the reward grows, but from where it is spent: per-dimension competition spreads improvement across dimensions instead of concentrating it in densely covered ones.

\begin{figure*}[t]
  \centering
  \includegraphics[width=\textwidth]{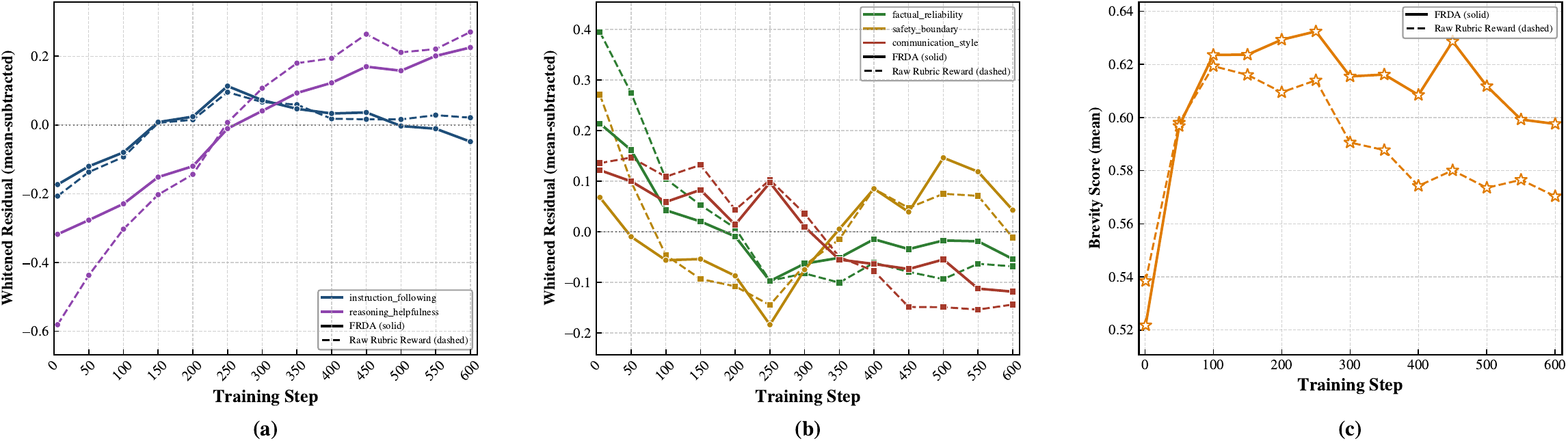}
  \caption{
    Dimension-level training dynamics on Chat with Qwen3-4B.
    (a) Whitened residuals for instruction following and reasoning
    helpfulness.
    (b) Whitened residuals for factual reliability, safety boundary,
    and communication style.
    (c) Scores on conciseness-related rubric criteria.
    Solid lines denote FRDA; dashed lines denote raw rubric reward.
  }
  \label{fig:figure33}
\end{figure*}

\noindent\textbf{Reduced Intra-Rubric Reward Divergence.}
The reward-growth argument above only speaks to the aggregated signal; we next check whether FRDA also changes the dimension-level dynamics that Figure~\ref{fig:figure12}(b) identifies as the source of rubric hacking.
Figures~\ref{fig:figure33}(a) and~\ref{fig:figure33}(b) plot each Chat dimension's Cholesky-whitened residual (Section~\ref{sec:method_latent}) for FRDA (solid) vs.\ raw rubric reward (dashed).
For dimensions trending downward under raw rubric reward, FRDA's decline is visibly shallower, and for those trending upward, the rise is similarly damped; both curves diverge increasingly from step 400 onward, with FRDA staying closer to its step-400 level.
This flattening coincides exactly with the window used for w-mean and $\Delta$ in Table~\ref{tab:main_results}, consistent with a single mechanism: by preventing any one dimension's residual from running away, FRDA keeps the policy from over-fitting a subset of criteria at the expense of others, rather than a general slowdown.

\noindent\textbf{Better Preservation of Conciseness.}
The dimension-level findings also appear in explicit
conciseness criteria.
Figure~\ref{fig:figure33}(c) shows that FRDA maintains higher
brevity scores during later training.
Table~\ref{tab:token_length} shows the corresponding difference
in response length: FRDA produces shorter responses at every
evaluated checkpoint from steps 400--600.
At step 600, the average length is 4{,}496 under FRDA,
compared with 5{,}180 under raw rubric reward.
The higher conciseness scores and stronger downstream results
show that FRDA limits excessive response growth while
preserving quality.

\begin{table}[t]
  \centering
  \resizebox{\columnwidth}{!}{%
  \begin{tabular}{lcc}
    \hline
    \textbf{Step} & \textbf{Raw Rubric Reward} & \textbf{FRDA} \\
    \hline
    400 & 3744 & 3279 \\
    450 & 4640 & 3254 \\
    500 & 4364 & 3639 \\
    550 & 4671 & 4113 \\
    600 & 5180 & 4496 \\
    \hline
  \end{tabular}%
  }
  \caption{
    Average response length on Arena-Hard v2.0 for Qwen3-4B.
    FRDA produces shorter responses throughout the comparison window.
  }
  \label{tab:token_length}
\end{table}

\noindent\textbf{Ablation: Average Dimension Mean vs.\ FRDA.}
A seemingly simpler fix for dimension imbalance is to average the unnormalized dimension-level rewards $r_{d}(x, y)$ with equal weights, replacing the weighted mean in Eq.~\eqref{eq:rubric_decomp} with a plain average over covered dimensions; we call this ablation Average Dimension Mean (ADM) and train it on Chat under the same setup as FRDA. Table~\ref{tab:adm_ablation} shows ADM falls below even the raw-rubric-reward baseline ($\Delta=-2.65$): equal-weight averaging ignores how differently separated each dimension's raw score distribution is, so a sparsely covered or noisy dimension with a wide score range dominates the aggregate, merely trading the original imbalance for a new one favoring high-variance dimensions. FRDA avoids this by normalizing each dimension's advantage to zero mean and unit variance within the sampling group (Eq.~\eqref{eq:dim_adv}) before averaging, so balancing acts on the group-relative advantage rather than the raw reward. Mitigating rubric hacking thus requires balancing at the right statistical level; naive re-weighting can introduce a bias as harmful as the one it removes.

\begin{table}[t]
  \centering
  \resizebox{\columnwidth}{!}{%
  \begin{tabular}{lccc}
    \hline
    \textbf{Method} & Peak & w-mean & $\Delta$ \\
    \hline
    Raw Rubric Reward & 70.79 & 67.41 & -- \\
    ADM & 70.82 & 64.76 & $-2.65$ \\
    FRDA (Ours)
    & \textbf{74.25} & \textbf{72.46} & \textbf{+5.05} \\
    \hline
  \end{tabular}%
  }
  \caption{
    Aggregation ablation on Arena-Hard v2.0 with Qwen3-4B.
    ADM equally averages raw dimension rewards;
    FRDA normalizes each dimension before aggregation.
    Metrics follow Table~\ref{tab:main_results}.
  }
  \label{tab:adm_ablation}
\end{table}

\subsection{Effectiveness of Residual Semantic Dimension Completion}
\label{sec:experiment_latent}

We evaluate RSDC on Chat using
Skywork-Reward-V2-Llama-3.1-8B as the holistic preference
reward model.
This setting provides a direct comparison between rubric rewards,
preference rewards, and their combination.

We compare five training variants:
raw rubric reward;
OP, which uses only the preference reward $r_{\mathrm{pref}}$;
MR, which sums the aggregate rubric and preference rewards;
FRDA+OP, which adds the preference reward as another dimension
in FRDA without whitening;
and RSDC, which applies Cholesky whitening to the joint
advantage vector before aggregation, with the preference
dimension placed last.
All variants use Qwen3-4B~\citep{qwen3} as policy model and are evaluated on Arena-Hard v2.0~\citep{areanahard}.

\begin{table}[t]
  \centering
  \resizebox{\columnwidth}{!}{%
  \begin{tabular}{lccc}
    \hline
    \textbf{Method} & Peak & w-mean & $\Delta$ \\
    \hline
    Raw Rubric Reward & 70.79 & 67.41 & -- \\
    OP & 78.03 & 77.08 & -- \\
    MR & 79.00 & 78.12 & +1.04 \\
    FRDA+OP & 79.03 & 78.13 & +1.05 \\
    RSDC (Ours)
    & \textbf{81.20} & \textbf{79.58} & \textbf{+2.50} \\
    \hline
  \end{tabular}%
  }
  \caption{
    Rubric--preference reward comparisons on Arena-Hard v2.0
    with Qwen3-4B.
    OP uses preference reward alone; MR sums rubric and preference
    rewards; FRDA+OP aggregates their dimension-wise normalized
    advantages; RSDC additionally applies Cholesky whitening.
    Peak and w-mean follow Table~\ref{tab:main_results}, while
    $\Delta$ is the w-mean difference relative to OP, the stronger
    single-reward baseline.
  }
  \label{tab:latent_ablation}
\end{table}

\noindent\textbf{Holistic Preferences Provide a Strong Additional Signal.}
Table~\ref{tab:latent_ablation} shows that OP alone achieves
a w-mean of 77.08, exceeding raw rubric reward by 9.67 points.
Combining the two signals with MR or FRDA+OP raises w-mean
to 78.12 and 78.13, respectively.
Both combinations outperform either reward source alone,
but their gains over OP are limited to approximately one point.

\noindent\textbf{RSDC Improves Rubric--Preference Integration.}
RSDC achieves the highest peak and w-mean, at 81.20 and 79.58.
Its w-mean exceeds OP by 2.50 points, MR by 1.46 points,
and FRDA+OP by 1.45 points.
By placing the preference dimension last, RSDC removes the
component linearly explained by the covered rubric dimensions
from its contribution to the aggregate advantage.
The improvement over FRDA+OP demonstrates the benefit of
whitening when combining these reward signals and supports
RSDC as an effective way to integrate rubric and holistic
preference feedback.

\nocite{Ando2005,andrew2007scalable,rasooli-tetrault-2015}

\section{Conclusion}
\label{sec:conclusion}

By decomposing rubric rewards into semantic dimensions, we identify Intra-Rubric Reward Divergence, a form of rubric hacking in which aggregate rubric reward keeps rising while specific sub-dimensions decline, and trace it to two causes: dimension imbalance and dimension absence. To address them, we propose Fair-Race Dimension Aggregation (FRDA), which normalizes and aggregates rubric sub-dimensions with equal weights so generations compete separately within each dimension, mitigating the dominance of densely covered dimensions over the reward signal, and Residual Semantic Dimension Completion (RSDC), which uses Cholesky whitening to extract from a holistic preference signal the latent dimensions not covered by the rubric. Experiments on chat, science, and medical domains show that FRDA effectively mitigates rubric hacking when the rubric training set has imbalanced dimension coverage and performs comparably to Dropout when it is balanced, while RSDC outperforms both single-reward training and naive multi-reward aggregation.

\section*{Limitations}
\label{sec:limitations}

Our analysis relies on a manual, coarse-grained partition of rubric criteria into semantic dimensions per domain, which requires domain-knowledgeable annotators and may vary across annotators; we do not pursue a fine-grained or universally agreed-upon decomposition.

Because specialized preference reward models for science and medicine are not readily available off the shelf, we validate RSDC only on chat, where a general-purpose conversational reward model exists; extending it to professional domains is left to future work.

%

\bibliography{custom}

\appendix

%

\end{document}